\documentclass[11pt]{article}

\usepackage{cql_style}
\usepackage[utf8]{inputenc}
\usepackage[T1]{fontenc}
\usepackage{lmodern}
\usepackage{amsmath, amssymb, amsfonts}
\usepackage{graphicx}
\graphicspath{{figures/}}
\usepackage{booktabs}
\usepackage{url}
\usepackage{nicefrac}
\usepackage{microtype}
\usepackage{algorithm}
\usepackage{algorithmic}
\usepackage{multirow}
\usepackage{makecell}
\usepackage{caption}
\usepackage{subcaption}
\usepackage{float}
\usepackage[section]{placeins}
\usepackage[numbers, compress]{natbib}

\title{Hubble: An LLM-Driven Agentic Framework \\ for Safe, Diverse, and Reproducible Alpha Factor Discovery}

\renewcommand{\thefootnote}{\fnsymbol{footnote}}

\author{%
  Runze Shi\footnotemark[1] \quad
  Shengyu Yan \quad
  Yuecheng Cai \quad
  Chenxi Lv \\
  \\
  Celestial Quant Lab, Canada \\
  The University of British Columbia, Canada \\
}

\date{March 2026}

\begin{document}

\maketitle
\footnotetext[1]{Corresponding author: \texttt{runze.shi@ubc.ca}}
\renewcommand{\thefootnote}{\arabic{footnote}}

\begin{abstract}
Automated alpha discovery is difficult because the search space of formulaic factors is combinatorial, the signal-to-noise ratio in daily equity data is low, and unconstrained program generation is operationally unsafe.
We present \textbf{Hubble}, an agentic factor mining framework that combines large language models (LLMs) with a domain-specific operator language, an abstract syntax tree (AST) execution sandbox, a dual-channel retrieval-augmented generation (RAG) module, and a family-aware selection mechanism.
Instead of treating the LLM as an unconstrained code generator, Hubble restricts generation to interpretable operator trees, evaluates every candidate through a deterministic cross-sectional pipeline, and feeds back both top formulas and structured family-level diagnostics to subsequent rounds.
The current system additionally introduces positive/negative RAG, formula-similarity penalties, standardized multi-metric scoring, dual reporting of RankIC and Pearson IC, and persistent diagnostics artifacts for post-hoc research analysis.
On a U.S. equity universe of roughly 500 stocks, our main run evaluates 104 valid candidates across three rounds with zero runtime crashes and discovers a top set dominated by \emph{range}, \emph{volatility}, and \emph{trend} families rather than crowded volume-only motifs.
We then fix the resulting top-$5$ factors and validate them on a held-out period from 2025-06-01 to 2026-03-13.
In this out-of-sample window, the two \emph{range} factors and two \emph{volatility} factors remain positive and several achieve HAC-significant Pearson IC and long-short evidence, whereas the weakest in-sample \emph{trend} factor decays materially.
These results suggest that safe LLM-guided search can be upgraded from a syntax-compliant generator into a reproducible alpha-research workflow that jointly optimizes validity, diversity, interpretability, and family-level generalization.

\vspace{1em}
\end{abstract}

\section{Introduction}

The search for predictive alpha factors remains a central task in quantitative equity investing~\citep{grinold2000active, qian2007quantitative}.
In practice, however, useful signals are rare, financial data are noisy, and the hypothesis space of symbolic formulas grows combinatorially.
Traditional automated discovery methods such as genetic programming can explore large search spaces efficiently~\citep{koza1992genetic}, but they often produce brittle expressions that are difficult to interpret, difficult to operationalize, and prone to overfitting.
On the other hand, hand-crafted libraries of formulaic alphas demonstrate the value of composable and interpretable expressions, but they remain labor-intensive to construct and extend~\citep{kakushadze2016alphafactory}.

Large language models (LLMs) offer a different search primitive: they can propose structured hypotheses conditioned on operator documentation, prior successful motifs, and explicit error feedback.
Yet a naive use of LLMs for formula mining introduces several failure modes at once: unsafe code execution, semantically invalid expressions, repeated rediscovery of crowded templates, and convergence toward a narrow set of high-coverage but low-novelty signals.
Recent work on LLM-based alpha mining has begun to address automated generation~\citep{alphagpt2023, alphaagent2025, factorminer2026}, but less attention has been paid to the joint requirements of execution safety, reproducibility, factor-family diversity, and research-ready diagnostics.

This paper presents the current version of \textbf{Hubble}, an agentic alpha mining framework built around four design principles:
\begin{enumerate}
    \item \textbf{Safe generation}: all candidates are constrained to a curated domain-specific language (DSL) and validated by a three-layer AST sandbox before execution.
    \item \textbf{Structured retrieval}: positive RAG supplies representative mechanisms to explore under-covered themes, while negative RAG explicitly discourages crowded templates.
    \item \textbf{Family-aware selection}: scoring and top-$k$ selection incorporate crowding, similarity, and factor-family concentration rather than rewarding raw predictive statistics alone.
    \item \textbf{Research-ready outputs}: every evaluated factor is archived with RankIC, Pearson IC, bucket returns, long-short spread, turnover, coverage, and formula complexity diagnostics.
\end{enumerate}

Compared with the earlier prototype, the current system is not merely safer or more convenient.
It changes the effective search behavior.
In the main run analyzed in this paper, the final top factors are no longer dominated by simple volume-normalization motifs; instead, the discovered factors span \emph{range}, \emph{volatility}, and \emph{price-trend} families.
This shift is important because it indicates that the framework is not only syntactically valid, but also capable of controlling diversity and avoiding over-crowded areas of the formula space.

Our contributions are therefore threefold.
First, we provide an end-to-end agentic framework for safe formula mining with deterministic storage, checkpointing, and structured diagnostics.
Second, we introduce a dual-channel RAG design and family-aware scoring layer that directly shape both exploration and final ranking.
Third, we demonstrate on a large daily U.S. equity panel that the framework can discover interpretable factors with coherent statistical and trading evidence, while remaining stable across multiple LLM backends.

\section{Related Work}

\paragraph{Automated factor discovery.}
Symbolic regression and genetic programming have long served as natural baselines for automated alpha discovery~\citep{koza1992genetic}.
Qlib and related platforms provide industrial-strength research infrastructure, but usually operate over predefined factor pools rather than open-ended generation~\citep{yang2020qlib}.
Formulaic alpha libraries such as the 101 alphas of \citet{kakushadze2016alphafactory} remain influential because they balance interpretability and composability, although their construction is still largely expert-driven.

\paragraph{LLM-driven alpha mining.}
Recent work has explored using LLMs as alpha generators, co-designers, or memory-augmented mining agents~\citep{alphagpt2023, alphaagent2025, factorminer2026}.
These systems highlight the promise of language-guided search, but they often emphasize generation quality more than execution safety, research reproducibility, or explicit control over factor crowding and diversity.
Our focus is narrower and more engineering-oriented: Hubble treats the LLM as a constrained hypothesis engine inside a deterministic evaluation loop.

\paragraph{LLM-guided search and reasoning.}
Beyond finance, LLMs have shown value as search heuristics for structured reasoning and program discovery~\citep{openai2023gpt4, wei2022chain, yao2023tree, romera2024mathematical}.
The closest analogy to our setting is program search with structured feedback: rather than directly emitting final solutions, the model proposes candidate programs that are subsequently filtered, scored, and iteratively refined.
Hubble applies this pattern to alpha mining while adding domain-specific control over DSL safety, factor-family coverage, and crowded-pattern avoidance.

\section{Methodology}

\begin{figure}[htbp]
\centering
\includegraphics[width=\textwidth]{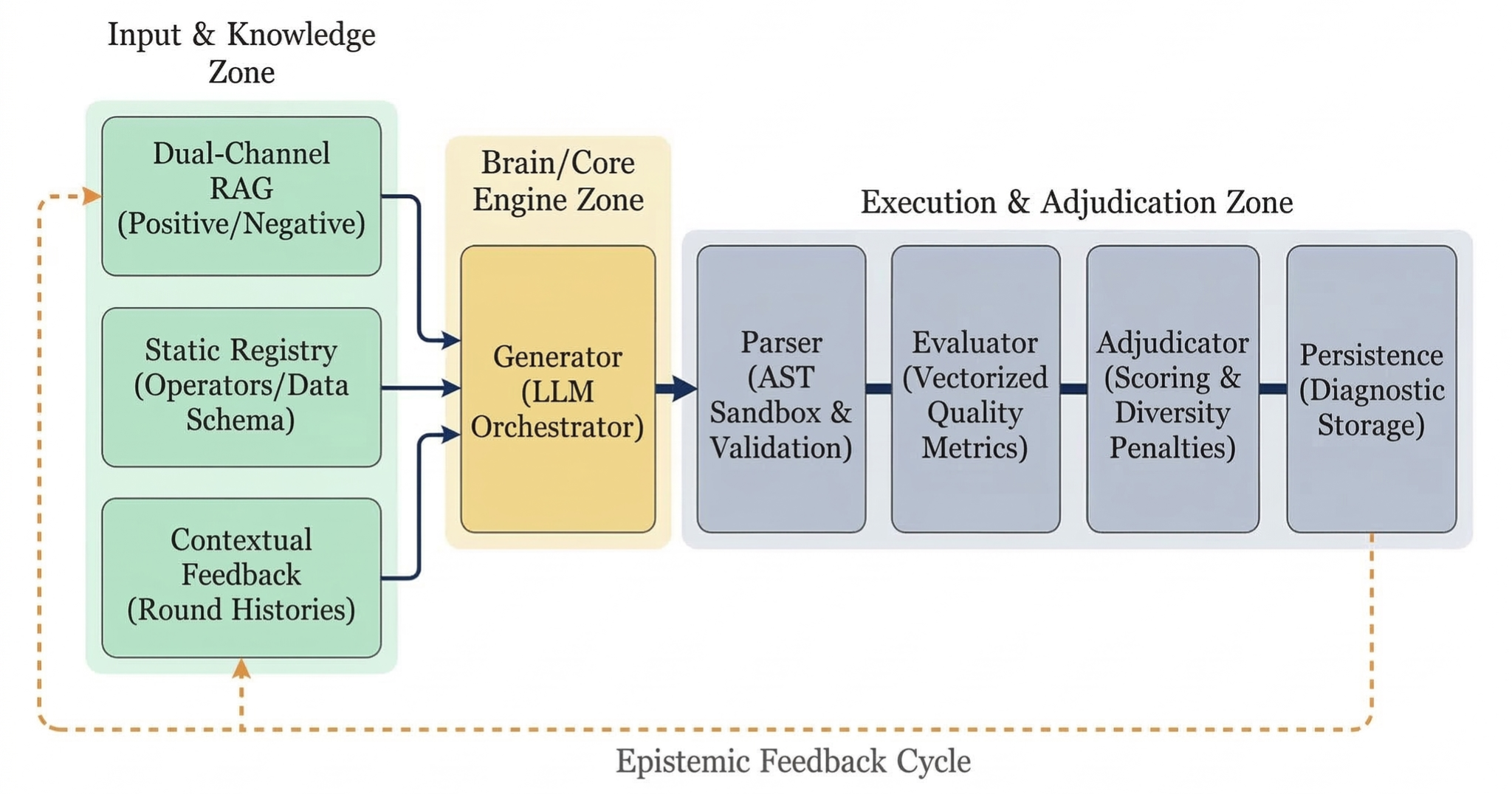}
\caption{Hubble agentic search loop. Positive and negative RAG references guide the generator; scored results and family diagnostics are fed back each round.}
\label{fig:architecture}
\end{figure}

The current Hubble pipeline is a closed-loop mining system composed of a DSL-constrained generator, an AST validation sandbox, a deterministic evaluation engine, a dual-channel RAG module, and a family-aware scorer.
Figure~\ref{fig:architecture} illustrates the modular agentic loop of the Hubble closed-loop factor mining pipeline.
The system consists of five core components: a DSL-constrained generator, a retrieval-enhanced RAG module, a safety parser with AST validation, a deterministic evaluation engine, and a weighted scoring module with archival feedback.

The Generator produces candidate formulas using an LLM under a constrained operator registry and dynamic prompt construction.
To guide exploration and avoid redundant factor discovery, the generator is augmented by a dual-channel RAG module, which retrieves metadata-aware samples from a factor corpus and provides both positive and negative examples.

All generated formulas are passed through a Parser safety sandbox, which enforces syntactic and semantic constraints via AST validation.
This sandbox ensures safe execution by disallowing arbitrary code execution (``exec-free''), while enforcing limits on expression depth, node count, variable alignment, and operator signatures.

Valid formulas are then executed by the evaluation engine, which performs data cleaning, alignment, and statistical quality checks.
Multiple metrics are computed, including rank-based information coefficient (RankIC), Pearson IC, their respective annualized information ratios (RankICIR and ICIR), turnover, bucket returns, and factor coverage.

The resulting metrics are aggregated by a weighted scoring module, which applies reward and penalty mechanisms based on predictive power, stability, turnover, and ecological diversity across factor families.
Finally, all results are archived in the storage and reporting layer, which records factor reports and compiles feedback signals used to guide subsequent search rounds, forming a fully closed-loop discovery system.

\subsection{DSL and AST Sandbox}

Candidate formulas are written in a restricted domain-specific language over registered operators and raw data fields.
The primitive inputs are daily panel series such as \texttt{OPEN}, \texttt{HIGH}, \texttt{LOW}, \texttt{CLOSE}, \texttt{VOLUME}, and \texttt{VWAP}; the operator registry contains arithmetic, time-series, cross-sectional, and logical operators.
Examples include \texttt{TS\_SMA}, \texttt{TS\_STD}, \texttt{TS\_VAR}, \texttt{TS\_LOGRET}, \texttt{CS\_RANK}, \texttt{CS\_ZSCORE}, and \texttt{IF}.

Each candidate expression is parsed into a Python abstract syntax tree and validated through three layers:
\begin{enumerate}
    \item \textbf{Structural security}: only a safe whitelist of AST node types is allowed.
    \item \textbf{Complexity control}: expressions exceeding preset depth or node-count limits are rejected.
    \item \textbf{Semantic validity}: operator names, arities, and variable names must all belong to the registered DSL.
\end{enumerate}
This design prevents arbitrary code execution while preserving a sufficiently expressive formula language for factor discovery.

\begin{figure}[htbp]
\centering
\includegraphics[width=0.6\textwidth]{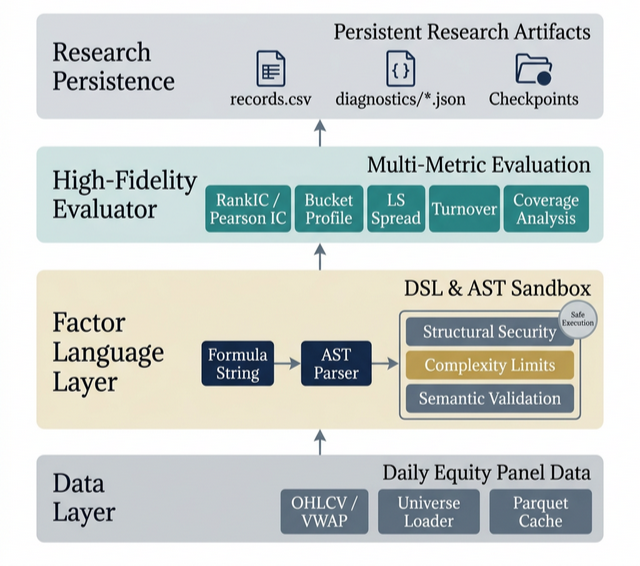}
\caption{Hubble infrastructure and safety stack, integrating market data ingestion, AST-based sandboxing, and multi-metric evaluation.}
\label{fig:architecture_stack}
\end{figure}

\subsection{Agentic Generation with Dual-Channel RAG}

At round $r$, the LLM receives a prompt bundle composed of a static operator prefix and a dynamic suffix that contains generation instructions, diversity requirements, optional feedback from round $r-1$, and optional RAG references.
Unlike the earlier version of the system, the current prompt explicitly requires theme diversity across factor families such as \emph{price trend}, \emph{mean reversion}, \emph{volatility}, \emph{range}, \emph{liquidity-volume}, and \emph{price-volume interaction}.

The RAG layer is \emph{dual channel}.
Positive RAG retrieves representative formulas or distilled templates from a curated corpus to encourage exploration of under-covered mechanisms.
Negative RAG retrieves crowded or over-explored templates and presents them as \emph{avoid-like} references.
This design follows the broader retrieval-augmented generation paradigm, but specializes retrieval into exploration and avoidance channels rather than pure factual augmentation~\citep{lewis2020rag}.
Operationally, the prompt therefore contains both:
\begin{itemize}
    \item \textbf{positive references}: mechanisms worth learning from, and
    \item \textbf{negative references}: formulas or structural motifs that should not be trivially re-parameterized.
\end{itemize}
This is a materially different use of retrieval than random few-shot prompting.
It turns the reference corpus into a control signal for exploration rather than a passive source of examples.

\subsection{Evaluation Pipeline}

For every validated factor $\alpha$, Hubble computes a forward return label and evaluates the factor cross-sectionally on each date.
The system reports both rank-based and linear metrics:
\begin{align}
    \mathrm{RankIC}_t &= \rho_S\!\left(\alpha_{t,\cdot}, r^{(h)}_{t,\cdot}\right), \\
    \mathrm{IC}_t &= \rho_P\!\left(\alpha_{t,\cdot}, r^{(h)}_{t,\cdot}\right),
\end{align}
where $\rho_S$ is Spearman correlation and $\rho_P$ is Pearson correlation.
The aggregated metrics include mean, standard deviation, daily information ratio, and annualized information ratio for both RankIC and IC.

In addition, the evaluator computes:
\begin{itemize}
    \item \textbf{coverage} and \textbf{drop ratio} after aligning factors with forward labels,
    \item \textbf{bucket returns} across factor quantiles,
    \item \textbf{long-short return} as the average top-minus-bottom bucket spread,
    \item \textbf{turnover} measured both on the top-decile set and on rank dynamics,\footnote{Top-decile turnover is defined as the Jaccard distance between consecutive top-decile sets: $1 - |A \cap B|\,/\,|A \cup B|$, where $A$ and $B$ are the top-decile asset sets on adjacent dates.}
    \item \textbf{formula complexity} via AST-derived depth, node count, operator count, and window count.
\end{itemize}

For significance testing, we test whether the time-series mean of each IC and long-short series differs from zero using a Bartlett-kernel HAC variance estimator with automatic lag selection following \citet{newey1987hac}.
These diagnostics are saved to per-factor artifacts and subsequently reused by the reporting and plotting layers.
The paper therefore uses the same persisted research outputs that the runtime system uses, rather than re-computing ad hoc metrics offline.

\subsection{Standardized Scoring and Family-Aware Selection}

The current scoring layer no longer uses a raw linear combination of heterogeneous metrics.
Instead, each metric $m_j$ is first standardized through a fixed center-and-scale transform,
\begin{equation}
    \tilde{m}_j = \tanh\!\left(\frac{m_j - c_j}{s_j}\right),
\end{equation}
where $(c_j, s_j)$ are metric-specific normalization constants.
The base score is then computed as
\begin{equation}
    \mathrm{score}_{\mathrm{base}}(\alpha) = \sum_j w_j \tilde{m}_j(\alpha),
\end{equation}
with positive weights for predictive and economic metrics and negative weights for turnover or data loss.

This base score is further adjusted by metadata-aware penalties and bonuses:
\begin{equation}
    \mathrm{score}(\alpha) =
    \mathrm{score}_{\mathrm{base}}(\alpha)
    - \lambda_c P_{\mathrm{crowded}}(\alpha)
    - \lambda_s P_{\mathrm{similar}}(\alpha)
    - \lambda_f P_{\mathrm{family}}(\alpha)
    + \lambda_n B_{\mathrm{novelty}}(\alpha).
\end{equation}
Here $P_{\mathrm{crowded}}$ penalizes proximity to crowded negative-RAG templates, $P_{\mathrm{similar}}$ penalizes near-duplicates of already selected formulas, and $P_{\mathrm{family}}$ discourages excessive concentration within a single factor family.
$B_{\mathrm{novelty}}$ is a small reward for candidates that expand the currently selected family set.
In the implementation, penalty and bonus magnitudes are pre-scaled when constructing the factor metadata; the scoring function receives them as direct additive adjustments to the base score, so the $\lambda$ coefficients above represent a conceptual decomposition rather than explicit run-time multipliers.

Selection is therefore performed under both \emph{quality} and \emph{diversity} constraints.
This matters empirically: in earlier runs, candidate generation became more diverse but the final top-$k$ remained dominated by liquidity-volume formulas.
The present scoring and selection scheme was introduced specifically to transmit diversity from the candidate pool into the final winners.

\subsection{Feedback and Persistent Artifacts}

At the end of each round, the system compiles feedback consisting of:
\begin{itemize}
    \item top formulas and their scores,
    \item error summaries,
    \item top, over-explored, and under-explored factor families,
    \item positive and negative RAG sample identifiers,
    \item prompt and usage metadata for reproducibility.
\end{itemize}

Every run is stored in a structured directory with append-only records, JSON summaries, prompt/response snapshots, and diagnostics artifacts.
This storage design is part of the research contribution because it makes the mining process auditable and reproducible rather than merely executable.

\section{Experiments}

\subsection{Data, Configuration, and Main Runs}

We evaluate Hubble on a daily U.S. equity universe loaded from \texttt{sp500.txt}.
The discovery run uses a panel of 501 valid stocks over 840 trading days from January 1, 2022 to May 31, 2025.
The main mining experiment is configured with three mining rounds, batch size 20, top-$k=5$, feedback enabled, RAG enabled, and the OpenRouter-backed model \texttt{nvidia/nemotron-3-super-120b-a12b:free}.
For backend robustness, we additionally run the same pipeline configuration with the backend model switched to \texttt{openrouter/hunter-alpha}.

The main run uses data strictly through 2025-05-31 (840 trading days), so the OOS period from 2025-06-01 to 2026-03-13 (195 trading days) is entirely unseen during factor discovery.
The five discovered factors, their formulas, and parameters are fixed before the OOS window begins.
No reranking or formula regeneration is performed on the OOS slice.

\begin{table}[t]
\centering
\caption{Main experimental configuration.}
\label{tab:setup}
\begin{tabular}{ll}
\toprule
\textbf{Item} & \textbf{Setting} \\
\midrule
Universe & S\&P 500 ticker list (\texttt{sp500.txt}) \\
Discovery panel & 501 valid stocks \\
Discovery date range & 2022-01-01 to 2025-05-31 \\
Discovery trading days & 840 \\
OOS window & 2025-06-01 to 2026-03-13 \\
OOS trading days & 195 \\
Rounds & 3 \\
Candidates per round & 20 \\
Feedback round & Enabled \\
RAG & Enabled, positive + negative channels \\
Main model & \texttt{nvidia/nemotron-3-super-120b-a12b:free} \\
Robustness model & \texttt{openrouter/hunter-alpha} \\
Top-$k$ family cap & 2 per family \\
\bottomrule
\end{tabular}
\end{table}

\subsection{Overall Mining Results}

Table~\ref{tab:round_summary} reports the per-round throughput of the main run.
Each round consists of a primary generation batch of 20 candidates followed by a feedback batch of up to 20 additional candidates, so the number of evaluated candidates per round exceeds the nominal batch size of 20.
Across three rounds, Hubble processed 60 primary candidates and completed 104 full evaluations.
All 104 evaluated factors completed successfully; the only errors were 16 duplicate detections.
No parser crash, runtime crash, or storage failure occurred during the run.

\begin{table}[t]
\centering
\caption{Per-round summary for the main run.}
\label{tab:round_summary}
\begin{tabular}{lrrrrr}
\toprule
\textbf{Round} & \textbf{Candidates} & \textbf{Evaluated} & \textbf{OK} & \textbf{Errors} & \textbf{Best Score} \\
\midrule
R1 & 20 & 40 & 40 & 0  & 5.195 \\
R2 & 20 & 34 & 34 & 6  & 5.622 \\
R3 & 20 & 30 & 30 & 10 & 5.436 \\
\midrule
\textbf{Total} & \textbf{60} & \textbf{104} & \textbf{104} & \textbf{16} & \textbf{5.622} \\
\bottomrule
\end{tabular}
\end{table}

\begin{figure}[htbp]
\centering
\includegraphics[width=0.78\textwidth]{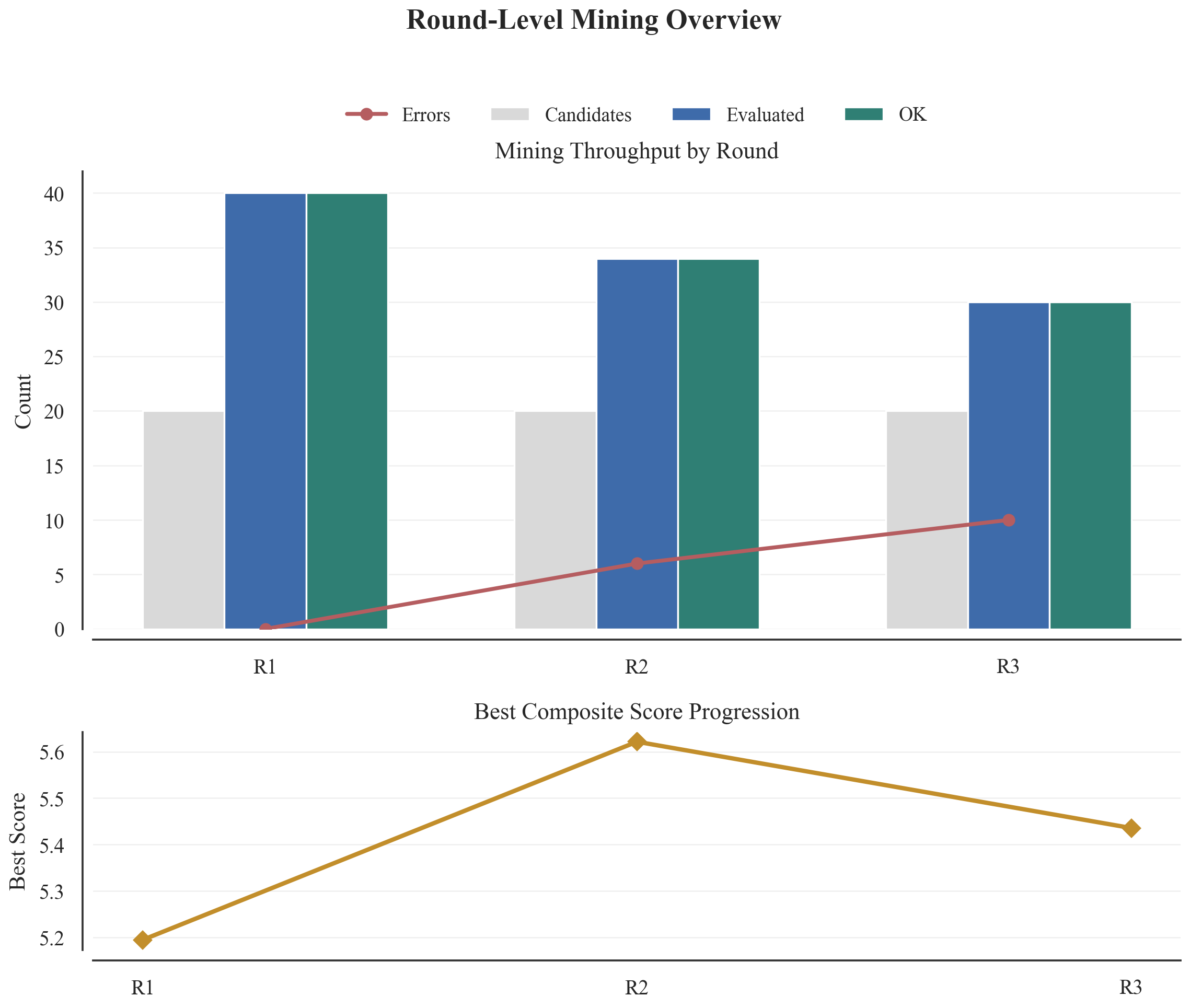}
\caption{Throughput and best-score progression across rounds for the main run.}
\label{fig:round_overview}
\end{figure}

Figure~\ref{fig:round_overview} shows that the feedback loop remains productive after the initial round.
The best score is achieved in Round~2 rather than Round~1, while the number of successful evaluations stays high.
This pattern is consistent with the intended behavior of the system: the first round broadens exploration, and later rounds refine themes without sacrificing runtime stability.

\subsection{In-Sample Factor Discovery}

Table~\ref{tab:factor_desc} summarizes the structural description of each discovered factor.
Table~\ref{tab:topk_is} reports the statistical performance on the discovery panel.
The selected set is concentrated in \emph{range}, \emph{volatility}, and \emph{price trend}, rather than collapsing into crowded liquidity-volume motifs.
Within the IS slice, the top factors are all directionally positive on both RankIC and Pearson IC, though the strongest HAC evidence is uneven across families.
This already suggests that the ranking signal is not purely driven by a single metric.

\begin{table}[t]
\centering
\caption{Structural descriptions of the top-$5$ factors.}
\label{tab:factor_desc}
\small
\begin{tabular}{llp{7.5cm}}
\toprule
\textbf{ID} & \textbf{Family} & \textbf{Structural Summary} \\
\midrule
Range-1     & range       & normalized intraday range with temporal smoothing \\
Range-2     & range       & normalized intraday range with smoothing and cross-sectional normalization \\
Volatility-1 & volatility & rolling variance of short-horizon log returns \\
Volatility-2 & volatility & rolling standard deviation of short-horizon log returns \\
Trend-1     & price trend & moving-average anchor relative to current price \\
\bottomrule
\end{tabular}
\end{table}

\begin{table}[t]
\centering
\caption{Top-$5$ factors from the main run, evaluated on the discovery panel (2022-01-01 to 2025-05-31).}
\label{tab:topk_is}
\footnotesize
\resizebox{\textwidth}{!}{%
\begin{tabular}{llrrrrrrr}
\toprule
\textbf{ID} & \textbf{Family} & \textbf{Score} & \textbf{RankIC} & \textbf{RankICIR (ann.)} & \textbf{IC} & \textbf{ICIR (ann.)} & \textbf{LS Ret.} & \textbf{Turnover} \\
\midrule
Range-1      & range       & 5.622 & 0.0059 & 0.350 & 0.0127 & 0.687 & 0.000810 & 0.113 \\
Range-2      & range       & 5.486 & 0.0057 & 0.342 & 0.0130 & 0.716 & 0.000724 & 0.161 \\
Volatility-1 & volatility  & 5.475 & 0.0057 & 0.368 & 0.0133 & 1.028 & 0.000546 & 0.174 \\
Volatility-2 & volatility  & 5.272 & 0.0047 & 0.297 & 0.0124 & 0.851 & 0.000587 & 0.137 \\
Trend-1      & price trend & 4.609 & 0.0115 & 0.900 & 0.0110 & 0.968 & 0.000350 & 0.368 \\
\bottomrule
\end{tabular}
}
\end{table}

The leading factor, denoted \textbf{Range-1}, is a smoothed normalized intraday range.
Its financial hypothesis is that sustained cross-sectional differences in normalized trading range proxy for latent information arrival, disagreement, and inventory pressure.
Stocks persistently exhibiting large normalized intraday ranges may be undergoing repricing or temporary demand imbalance that is not fully incorporated into next-period returns.

The strongest volatility-family factor, denoted \textbf{Volatility-1}, measures recent variance of daily log returns.
Its financial hypothesis is related to volatility clustering and conditional risk repricing: assets with elevated recent idiosyncratic activity may continue to experience cross-sectional repricing as information diffuses and investors update risk demands.

The trend-family representative, denoted \textbf{Trend-1}, compares a recent moving-average anchor to the current price.
Its financial interpretation is a relative dislocation-to-trend-anchor measure.
When used as a cross-sectional rank, it captures whether an asset is trading below or above a recent trend baseline, which can proxy for short-horizon mean reversion, trend continuation, or delayed adjustment depending on the cross-sectional context.

Taken together, the IS results indicate that Hubble is not merely rediscovering one crowded motif.
Rather, it recovers a compact family of interpretable mechanisms tied to price range, volatility state, and relative trend positioning.

\subsection{Out-of-Sample Validation}

Since the main run uses data strictly through 2025-05-31, the OOS period from 2025-06-01 to 2026-03-13 is entirely held out from factor discovery.
The five factors, their formulas, and parameters are fixed before the OOS window begins.
Table~\ref{tab:topk_oos} shows that the four higher-ranked factors remain directionally positive in OOS, while \textbf{Trend-1} is the only factor that clearly deteriorates.

\begin{table}[t]
\centering
\caption{Out-of-sample validation for the top-$5$ factors (2025-06-01 to 2026-03-13). Significance stars are based on HAC standard errors: * $p<0.05$, ** $p<0.01$, *** $p<0.001$.}
\label{tab:topk_oos}
\footnotesize
\resizebox{\textwidth}{!}{%
\begin{tabular}{llrrrrrrrrrr}
\toprule
\textbf{ID} & \textbf{Family} & \textbf{RankIC} & \textbf{RankICIR (ann.)} & \textbf{RankIC $t$ (HAC)} & \textbf{IC} & \textbf{ICIR (ann.)} & \textbf{IC $t$ (HAC)} & \textbf{LS Ret.} & \textbf{LS $t$ (HAC)} & \textbf{Turnover} \\
\midrule
Range-1      & range       & 0.0242 & 1.804 & 1.75    & 0.0391  & 2.788  & 2.98**  & 0.00205  & 2.17*   & 0.133 \\
Range-2      & range       & 0.0245 & 1.860 & 1.81    & 0.0385  & 2.797  & 3.01**  & 0.00209  & 2.31*   & 0.182 \\
Volatility-1 & volatility  & 0.0275 & 2.305 & 2.25*   & 0.0131  & 1.656  & 1.68    & 0.00200  & 2.60**  & 0.173 \\
Volatility-2 & volatility  & 0.0232 & 1.900 & 1.81    & 0.0204  & 2.251  & 2.47*   & 0.00189  & 2.27*   & 0.133 \\
Trend-1      & price trend & 0.0083 & 0.794 & 0.81    & $-$0.0034 & $-$0.380 & $-$0.39 & $-$0.00064 & $-$0.87 & 0.374 \\
\bottomrule
\end{tabular}
}
\end{table}

The most important pattern is family-specific.
Both \textbf{Range-1} and \textbf{Range-2} strengthen materially in OOS: their Pearson IC HAC $t$-statistics reach 2.98 and 3.01, respectively, and both retain significant long-short evidence.
\textbf{Volatility-1} also remains strong, achieving a RankIC HAC $t$-statistic of 2.25 and a long-short HAC $t$-statistic of 2.60.
\textbf{Volatility-2} is slightly weaker on rank-based evidence but remains supported by Pearson IC and long-short significance.
In contrast, \textbf{Trend-1}, which already had the lowest composite score in-sample, fails to generalize: its Pearson IC reverses sign and its long-short spread is no longer significant.

This differential OOS outcome is informative.
It suggests that the in-sample composite ranking is not arbitrary: the factor that the mining system ranked last is also the one that decays most clearly in holdout evaluation.
At the same time, the OOS window favors \emph{range} and \emph{volatility} families, so these results should be interpreted as preliminary holdout evidence rather than unconditional alpha claims.

\subsection{Visual Evidence}

\begin{figure}[htbp]
\centering
\includegraphics[width=0.88\textwidth]{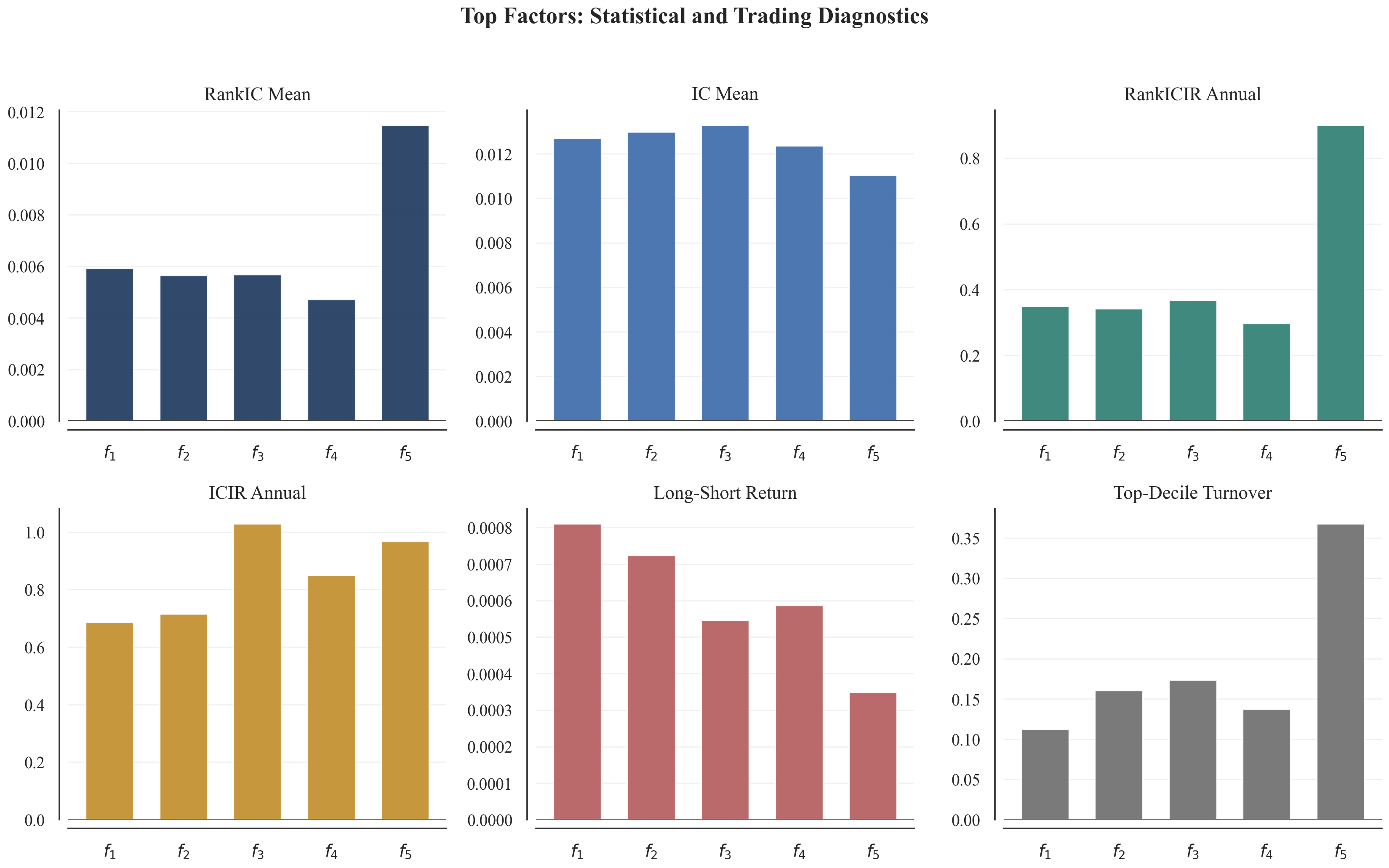}
\caption{In-sample statistical and trading diagnostics for the top-$5$ factors.}
\label{fig:is_metrics}
\end{figure}

\begin{figure}[htbp]
\centering
\includegraphics[width=0.76\textwidth]{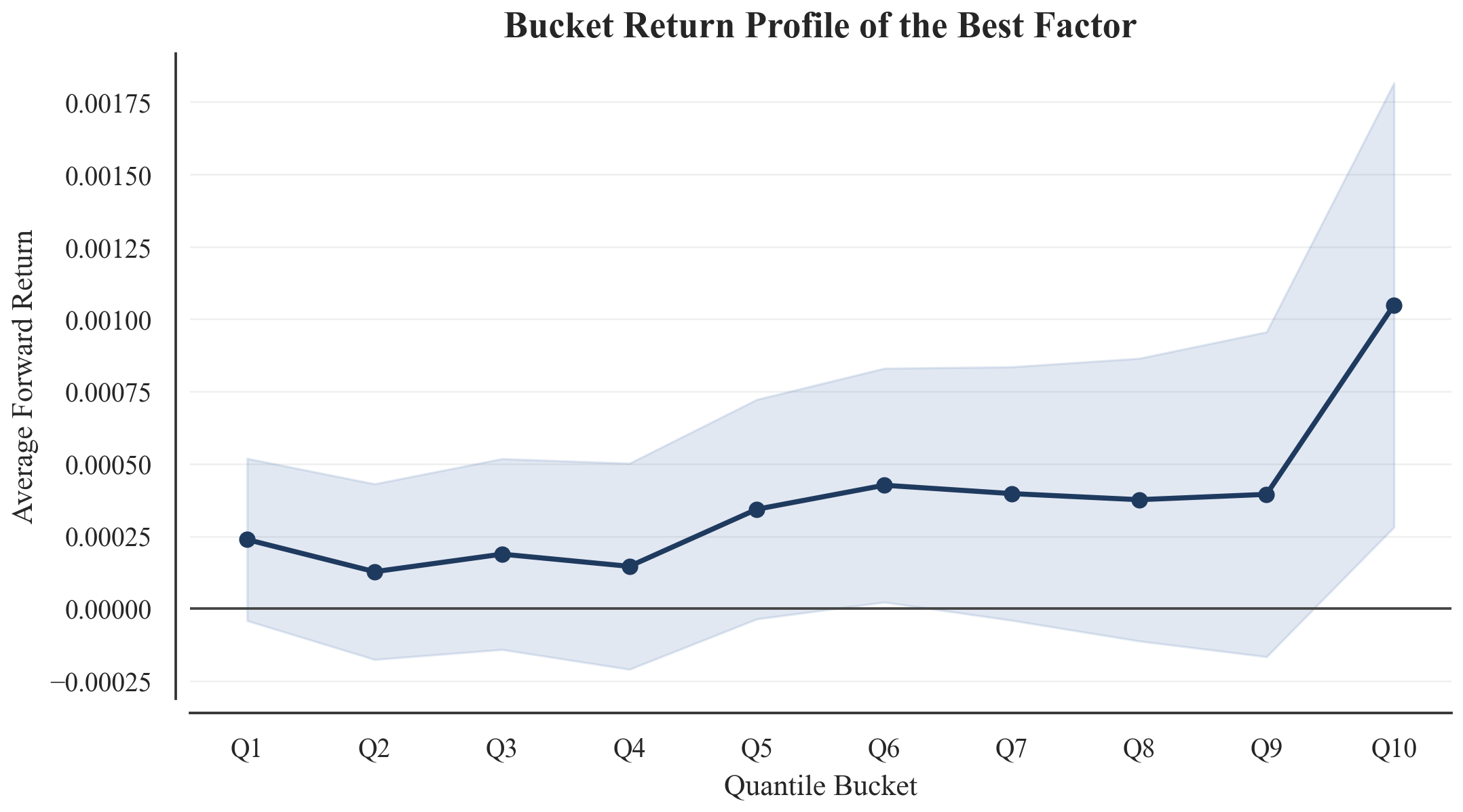}
\caption{In-sample bucket-return profile of the best factor.}
\label{fig:is_bucket}
\end{figure}

The visual evidence is organized as paired IS/OOS figures so that discovery and validation can be read together.
The IS diagnostics in Figure~\ref{fig:is_metrics} and Figure~\ref{fig:is_bucket} show that the top factors already form sensible cross-sectional structures before the holdout period begins.
The corresponding OOS figures then reveal which structures persist.

\begin{figure}[htbp]
\centering
\includegraphics[width=0.88\textwidth]{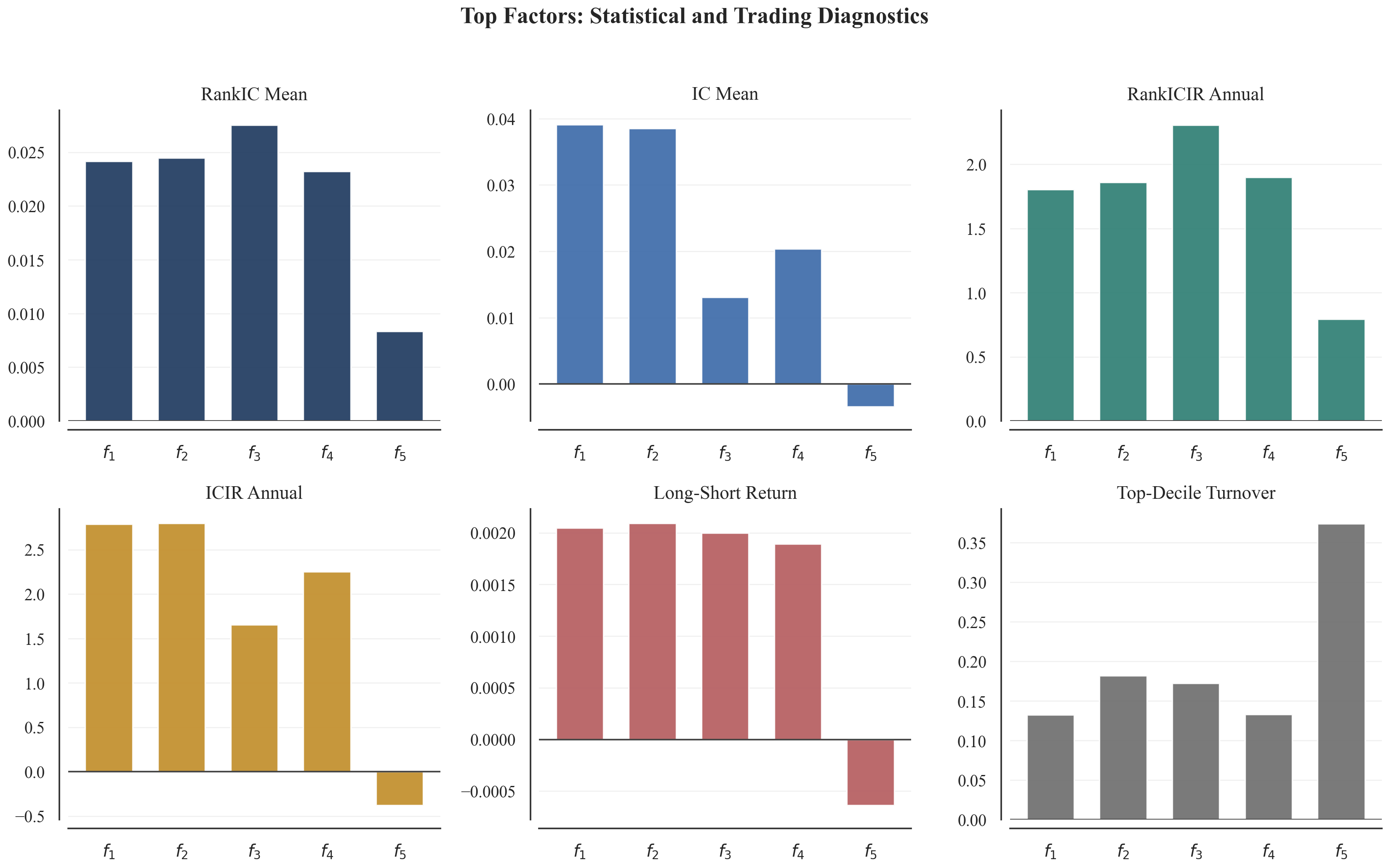}
\caption{Out-of-sample statistical and trading diagnostics for the top-$5$ factors.}
\label{fig:oos_metrics}
\end{figure}

\begin{figure}[htbp]
\centering
\includegraphics[width=0.76\textwidth]{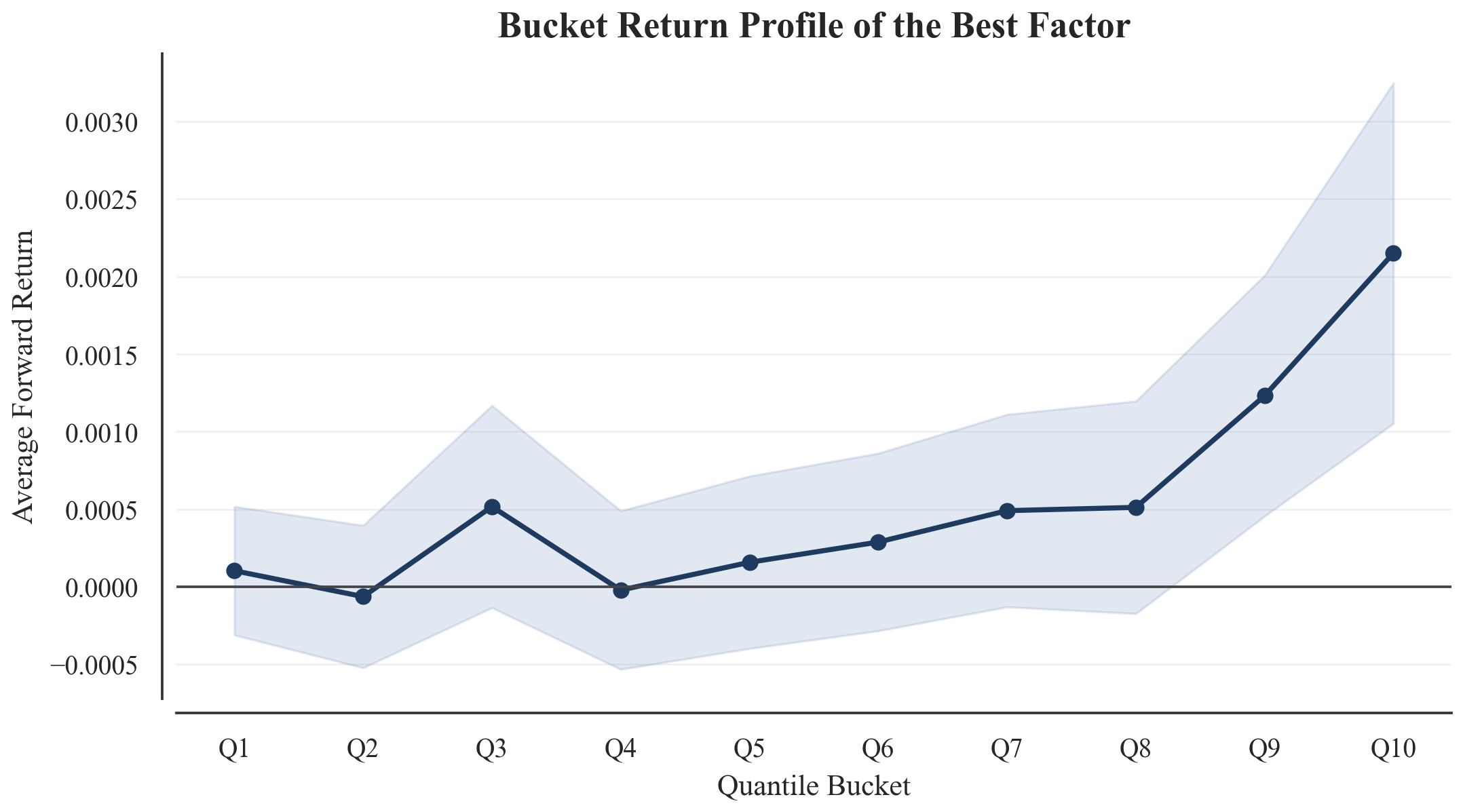}
\caption{Out-of-sample bucket-return profile of the best factor.}
\label{fig:oos_bucket}
\end{figure}

\begin{figure}[htbp]
\centering
\includegraphics[width=0.90\textwidth]{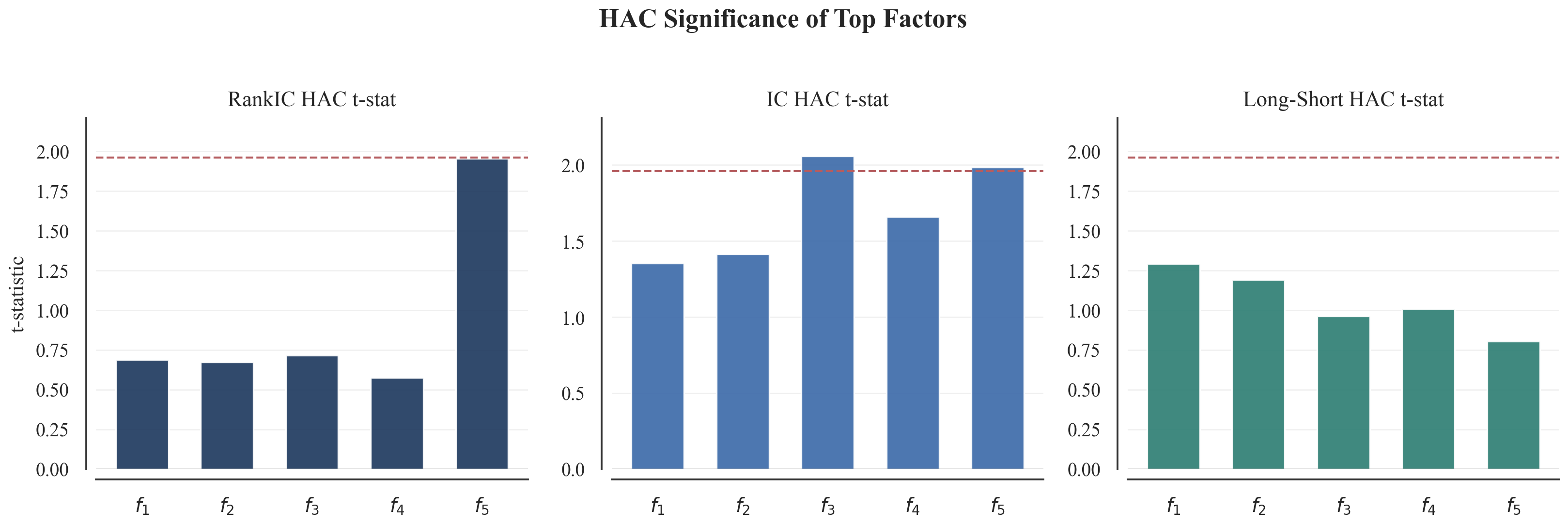}
\caption{HAC significance of the top-$5$ factors in-sample.}
\label{fig:is_sig}
\end{figure}

\begin{figure}[htbp]
\centering
\includegraphics[width=0.90\textwidth]{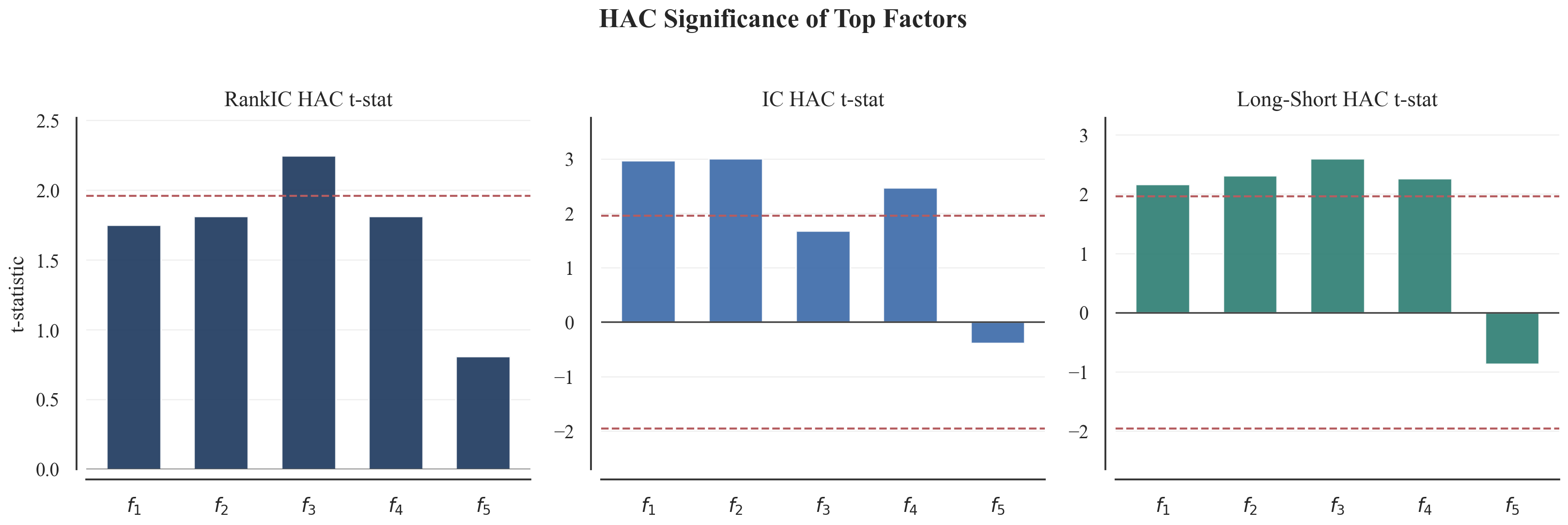}
\caption{HAC significance of the top-$5$ factors out-of-sample.}
\label{fig:oos_sig}
\end{figure}

\begin{figure}[htbp]
\centering
\includegraphics[width=0.90\textwidth]{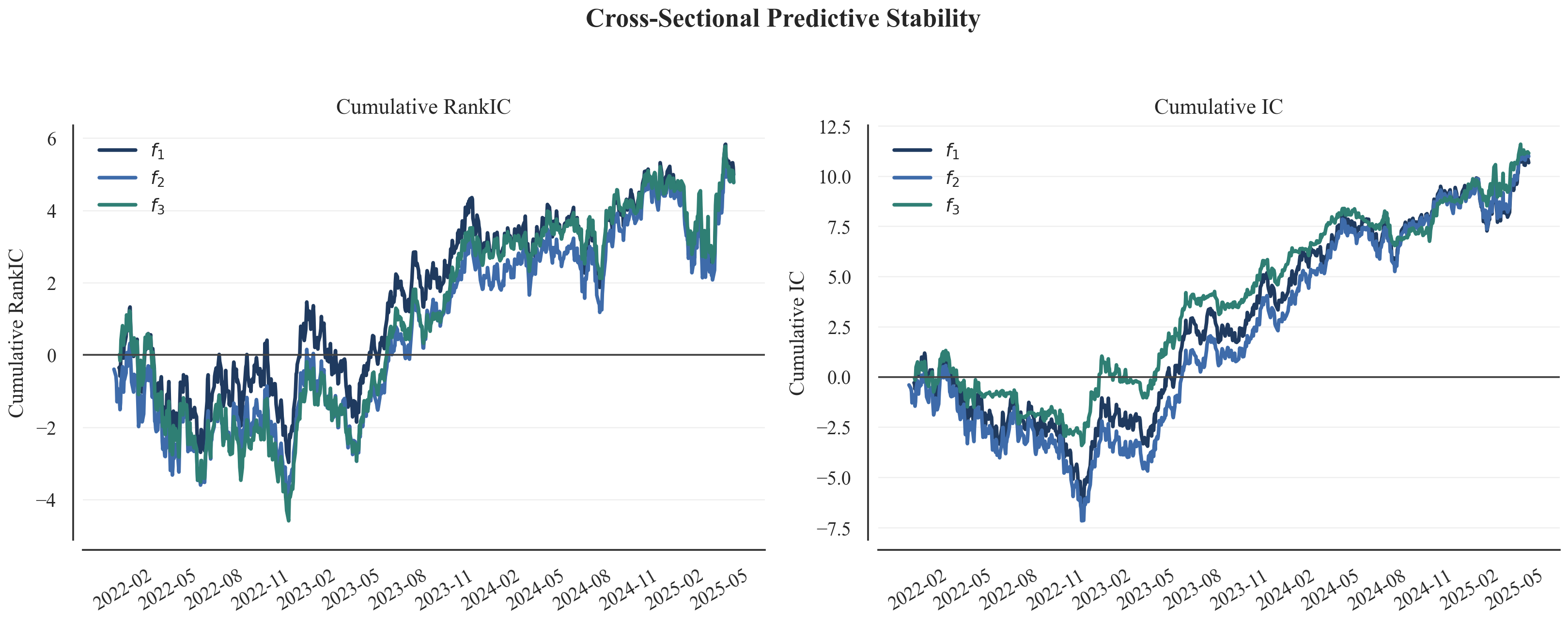}
\caption{Cumulative IC for the top-$3$ factors in-sample.}
\label{fig:is_ic_stability}
\end{figure}

\begin{figure}[htbp]
\centering
\includegraphics[width=0.90\textwidth]{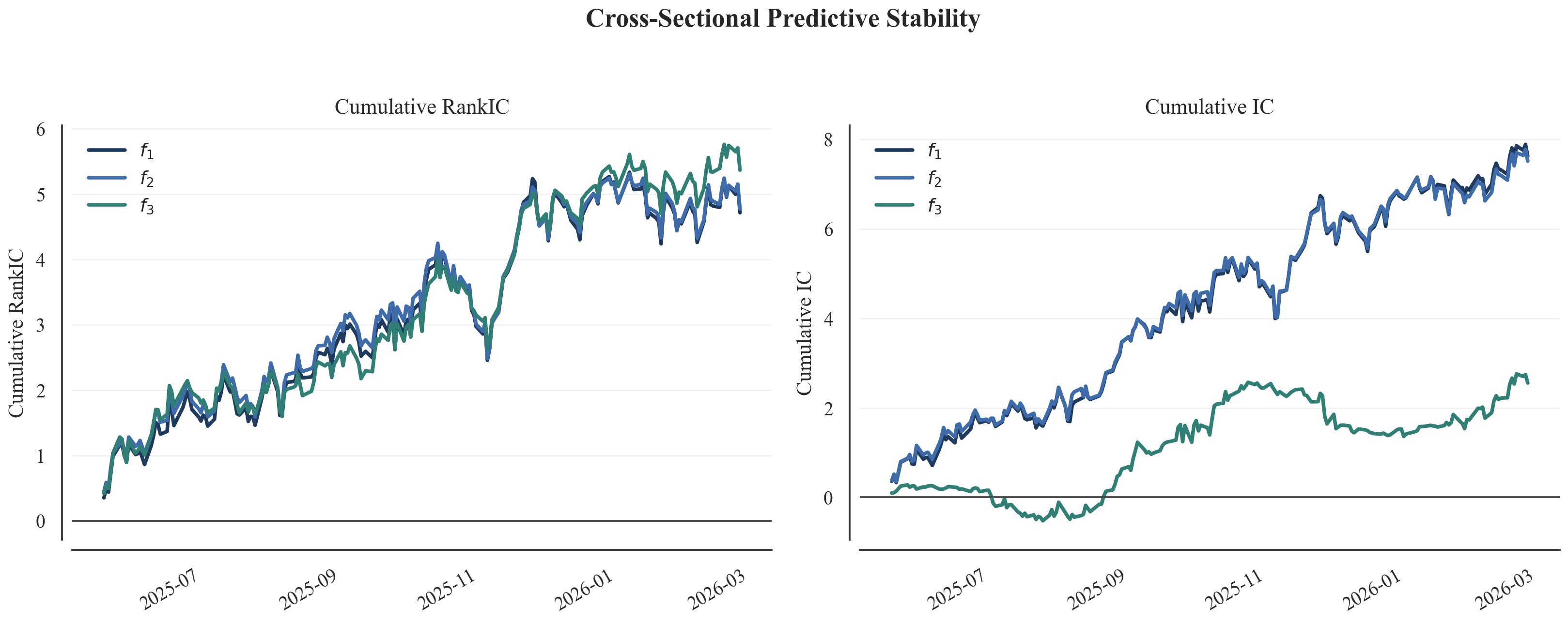}
\caption{Cumulative IC for the top-$3$ factors out-of-sample.}
\label{fig:oos_ic_stability}
\end{figure}

\begin{figure}[htbp]
\centering
\includegraphics[width=0.86\textwidth]{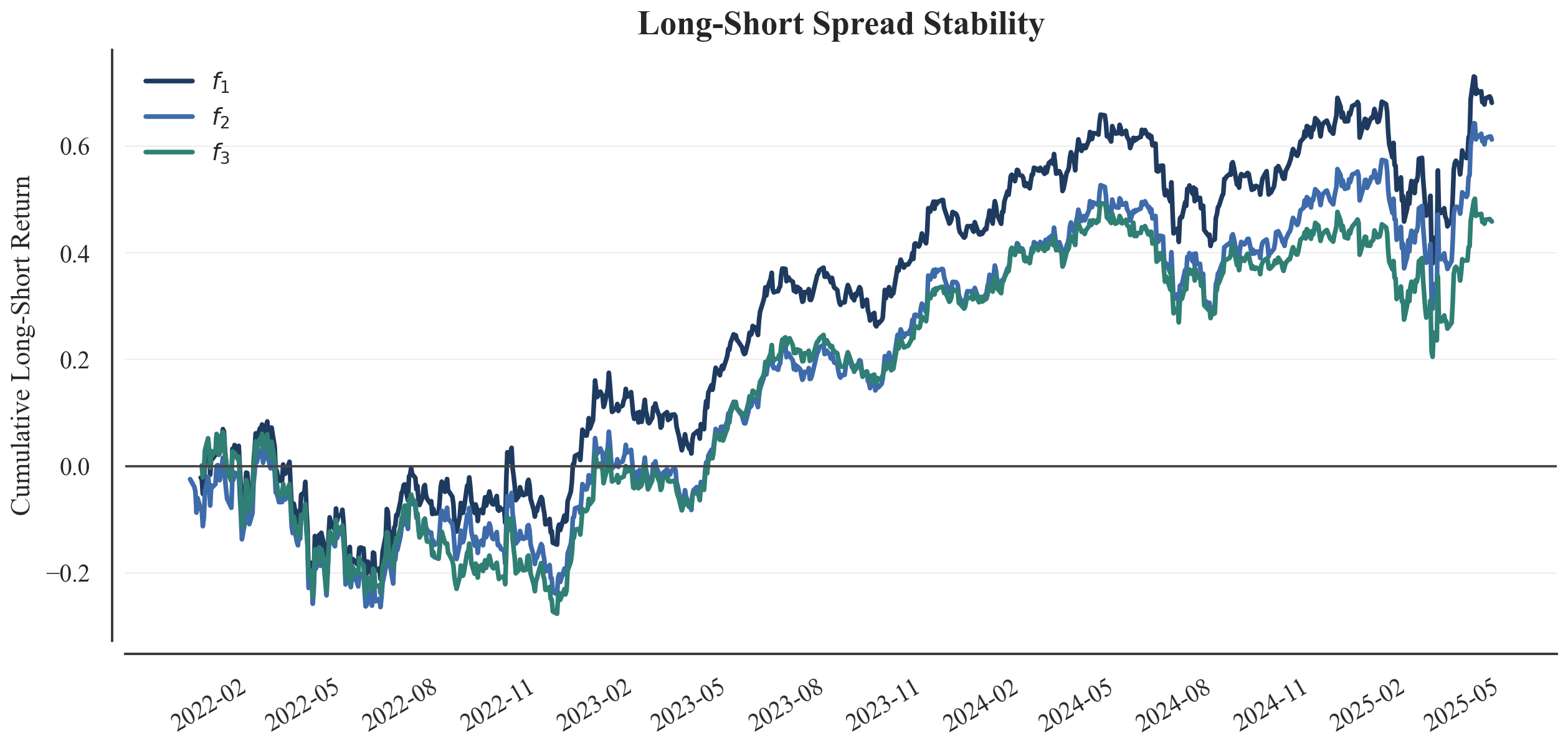}
\caption{Cumulative long-short spread of the top-$3$ factors in-sample.}
\label{fig:is_long_short}
\end{figure}

\begin{figure}[htbp]
\centering
\includegraphics[width=0.86\textwidth]{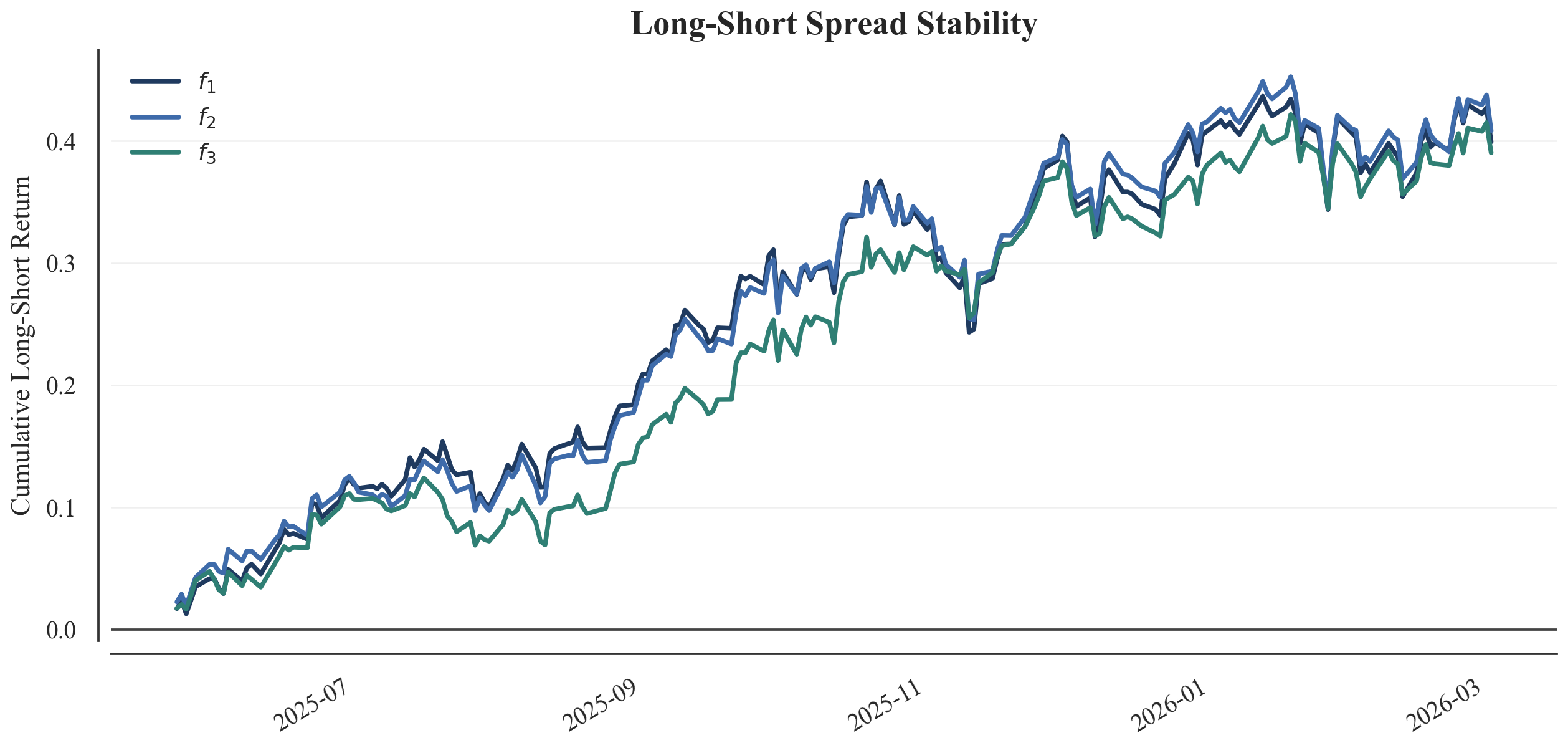}
\caption{Cumulative long-short spread of the top-$3$ factors out-of-sample.}
\label{fig:oos_long_short}
\end{figure}

\begin{figure}[htbp]
\centering
\includegraphics[width=0.82\textwidth]{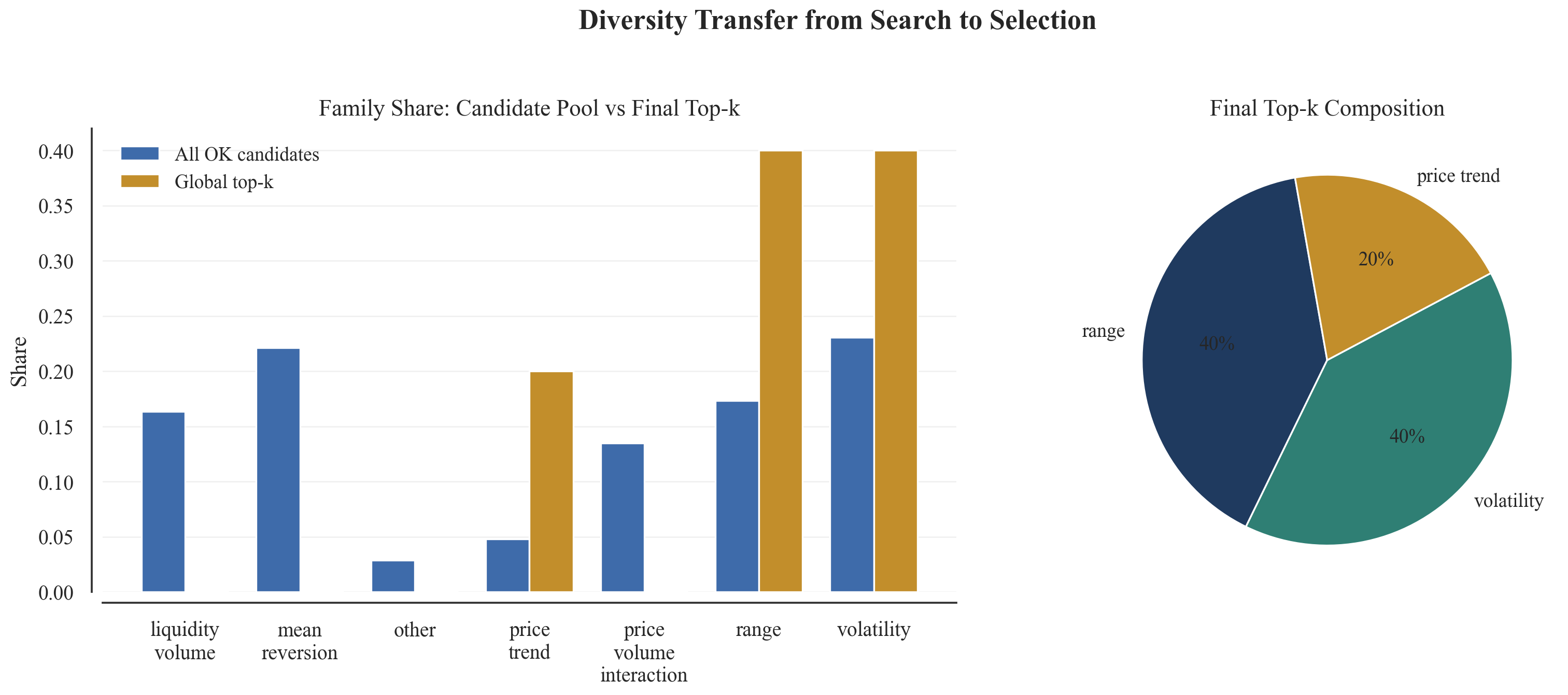}
\caption{Family composition of the main run: candidate pool versus final top-$k$ selection.}
\label{fig:family_diversity}
\end{figure}

Finally, diversity should be visible not only in the candidate pool but also in the selected winners.
Figure~\ref{fig:family_diversity} therefore compares the family mix of all successful candidates with that of the final top-$k$.
The key point is that the final winners remain distributed across range, volatility, and trend families, rather than collapsing back into a single crowded theme.

\subsection{Robustness Across LLM Backends}

To test whether the result is idiosyncratic to one backend, we compare the main run with a robustness run that uses \texttt{openrouter/hunter-alpha} under the same mining pipeline.
The robustness run is slightly weaker in peak score and includes 15 total errors (13 duplicate detections and 2 evaluation failures), but it recovers a qualitatively similar factor profile: top factors still come from the \emph{range}, \emph{volatility}, and \emph{price trend} families, with positive RankIC, positive Pearson IC, and meaningful long-short spreads.

\begin{table}[H]
\centering
\caption{Robustness comparison across two LLM backends.}
\label{tab:robustness}
\small
\begin{tabular}{p{2.8cm}p{3.8cm}p{3.8cm}}
\toprule
\textbf{Metric} & \textbf{Main Run} & \textbf{Robustness Run} \\
\midrule
Model & Nemotron-120B-free & Hunter-Alpha \\
Total evaluated & 104 & 107 \\
Total OK & 104 & 105 \\
Total errors & 16 & 15 \\
Best score & 5.622 & 5.460 \\
Top families & range, volatility, trend & range, volatility, trend \\
Best factor type & smoothed normalized intraday range & averaged normalized intraday range \\
\bottomrule
\end{tabular}
\end{table}

This robustness check is not intended to replace the OOS analysis.
Rather, it shows that the revised system design does not depend on a single backend to escape the earlier liquidity-volume collapse.
That is an important practical property for public deployment and reproducible research.

\paragraph{On formula disclosure.}
For a public arXiv version, we intentionally avoid disclosing exact factor formulas and hyper-parameters.
Instead, we report factor families, structural summaries, and finance-motivated hypotheses.
This preserves the scientific value of the paper by making the discovered mechanisms interpretable, while avoiding direct release of production-grade alpha specifications.

\section{Discussion}

The most important empirical finding is not the absolute magnitude of the composite score.
It is the fact that diversity constraints, negative RAG, and family-aware scoring changed the \emph{identity} of the winning factors and that this change survives a holdout check at the family level.
Earlier versions of the pipeline generated broader candidate pools but still selected mostly liquidity-volume variants at the end.
In contrast, the current system produces a final top set dominated by \emph{range}, \emph{volatility}, and one weaker \emph{trend} factor, and the OOS analysis shows that the two range factors and two volatility factors generalize more convincingly than the trend factor.

This shift matters for at least three reasons.
First, it reduces the risk that the framework is simply optimizing for high-coverage crowded templates.
Second, it yields more interpretable hypotheses that can be discussed in financial terms rather than only statistically.
Third, it makes the framework more useful as a research assistant: the archived diagnostics, bucket profiles, long-short curves, and HAC significance reports can be inspected directly when deciding whether a discovered factor deserves follow-up analysis.

The OOS results also suggest a second, subtler point: in-sample ranking does not perfectly determine cross-period robustness.
The factor with the lowest in-sample composite score, \textbf{Trend-1}, is also the one that fails most clearly out-of-sample.
However, the strongest OOS evidence concentrates in the range and volatility families rather than being distributed uniformly across all five winners.
This supports a family-level interpretation of generalization: the framework appears better at discovering stable \emph{mechanisms} than at guaranteeing that every top-ranked formula will generalize equally well.

At the same time, the absolute correlation magnitudes remain small, as is typical in daily cross-sectional equity prediction.
This should not be misread as weakness by itself; in this domain, even small cross-sectional predictive correlations can be meaningful if they are stable, diversified, and economically coherent.
That is why we report both IC-style measures and portfolio-style diagnostics, and why we interpret the OOS evidence as supportive but not definitive.

\section{Limitations}

\begin{itemize}
    \item \textbf{No neutralization.} The discovered factors are evaluated without market, sector, or style neutralization. The reported IC and long-short statistics may therefore incorporate systematic risk exposures---such as market beta or sector tilts---in addition to idiosyncratic alpha signals. This is a likely contributor to the elevated OOS IC magnitudes, particularly for range and volatility factors during the 2025--2026 window when broad market volatility was elevated. Neutralized factor evaluation remains an important direction for future work.

    \item \textbf{Single panel, no walk-forward.} The present analysis is based on a single U.S. daily equity panel and does not yet include a full walk-forward out-of-sample protocol. The OOS validation in this paper is informative, but it remains one temporal split rather than a rolling retraining study.

    \item \textbf{Short OOS window and regime risk.} The OOS window spans only 195 trading days and may partly reflect a regime favorable to range- and volatility-sensitive signals. The positive OOS results should therefore be interpreted as preliminary holdout evidence rather than unconditional alpha claims. This caution is consistent with the broader evidence that many published return predictors attenuate materially after discovery or fail to replicate cleanly across samples~\citep{mclean2016predictability,hou2020replicating,harvey2016backtesting}.

    \item \textbf{RAG corpus bias.} The RAG corpus still reflects curated historical examples and may retain residual thematic bias from earlier experiments.

    \item \textbf{Lightweight robustness analysis.} The robustness analysis across backends is intentionally lightweight; a more complete public version should include explicit ablations such as RAG on/off and standardized scoring on/off.

    \item \textbf{No transaction-cost evaluation.} Formula quality is assessed through statistical and simple portfolio diagnostics rather than a full transaction-cost-aware backtest.

    \item \textbf{LLM temporal leakage.} The LLM generator operates under a temporal isolation constraint that applies only at the data layer. The model's pretraining corpus necessarily includes financial literature and research published after the IS window begins, meaning the LLM may have internalized meta-knowledge about which factor families or structural motifs have been reported as effective in the subsequent period. The DSL constraint prevents direct encoding of historical return series, but cannot prevent the model from having absorbed documented findings about factor effectiveness. A fully rigorous temporal firewall would require retraining the generator on a strictly pre-discovery corpus, which is not currently feasible with publicly available models.

    \item \textbf{LLM narrative bias.} The LLM's generative fluency introduces a systematic narrative risk. Because language models are trained to produce coherent, domain-consistent explanations, every discovered factor arrives paired with a ready-made financial rationale. A spurious signal accompanied by a compelling mechanism story is structurally harder to reject than a transparently flawed one: the plausibility of the explanation lowers the researcher's critical threshold independent of the statistical evidence. Hubble's deterministic evaluation pipeline and held-out OOS validation partially offset this risk, but they cannot eliminate it. Practitioners should treat the LLM-generated financial hypotheses as prompts for further scrutiny, not as independent corroboration of a factor's validity.
\end{itemize}

\section{Conclusion}

We presented the current public version of Hubble, an agentic framework for safe and diverse alpha factor discovery.
The system combines a DSL-constrained LLM generator, an AST execution sandbox, dual-channel RAG, standardized multi-metric scoring, family-aware selection, and persistent diagnostics artifacts.
On a 501-stock U.S. equity panel, the framework achieved complete runtime stability in the main run while discovering interpretable factors from range, volatility, and trend families.
Under a holdout validation from 2025-06-01 to 2026-03-13, the strongest evidence remained concentrated in the range and volatility families, while the weakest trend factor decayed materially.

The main value of the current system is not that it replaces traditional alpha research.
Rather, it provides a reproducible research workflow in which LLMs propose structured hypotheses, deterministic components adjudicate validity and quality, and the resulting artifacts are immediately usable for financial interpretation and follow-up testing.
Future work should focus on explicit ablations, walk-forward validation, transaction-cost-aware evaluation, and continued refinement of the RAG corpus to improve novelty without sacrificing reliability.

\bibliographystyle{unsrtnat}
\bibliography{references}

\end{document}